\documentclass[letterpaper, 10 pt, conference]{ieeeconf}  

\IEEEoverridecommandlockouts                              

\usepackage{graphics} 
\usepackage{epsfig} 
\usepackage{mathptmx} 
\usepackage{times} 
\usepackage{amsmath} 
\usepackage{amssymb}  
\usepackage{hyperref} 

\usepackage{booktabs} 
\usepackage{multirow} 
\usepackage{siunitx}

\title{\LARGE \bf
Approximating High Dimensional Self-Motion Manifolds via Deep Generative Models
}

\author{
    Haitao Gao$^{1}$, Yang Song$^{2}$, and Liao Wu$^{1}$%
    \thanks{This work was supported by the Australian Research Council under Grant DP250102489.}%
    \thanks{$^{1}$Haitao Gao and Liao Wu are with the School of Mechanical and Manufacturing Engineering,
    University of New South Wales, Sydney, Australia
    {\tt\small haitao.gao@unsw.edu.au}.}%
    \thanks{$^{2}$Yang Song is with the School of Computer Science and Engineering,
    University of New South Wales, Sydney, Australia.}%
}

\begin{document}

\maketitle
\thispagestyle{empty}
\pagestyle{empty}

\begin{abstract}

Self-motion manifold (SMM) characterizes the geometric structure of the
infinite inverse kinematic solutions set of a redundant manipulator at a fixed end-effector pose,
and its efficient recovery underpins feasible and global optimal
motion planning. Existing methods such as null-space continuation and learning-based methods are formulated around the assumption
that an SMM is a curve, and do not extend to higher redundancy orders. We
instead adopt a probabilistic view: SMMs are the support of the conditional
posterior over configurations given a target pose, so that recovering it reduces
to sampling from a learned distribution and separating its disjoint components
by clustering. The formulation is independent of the manifold dimension and
requires no architectural change as the redundancy order grows. In this work, we demonstrate that our method can approximate 1-D SMMs with performance comparable to the latest null-space continuation and learning-based approach, and that it is the first method capable of approximating highly redundant 4-D SMMs in a 7R manipulator for position tasks. Project website: \href{https://github.com/accuracy-maker/high-dimenstional-self-motion-manifold-approximation}{https://github.com/accuracy-maker/high-dimenstional-self-motion-manifold-approximation}

\end{abstract}

\section{INTRODUCTION}
Robotic system design has progressively evolved toward achieving higher kinematic dexterity and enhanced flexibility in motion control~\cite{brady1982robot, lavalle2006planning}. The available degrees of freedom (DoF) have increased from classical 2R and 3R serial manipulators for planar motion, through industrial 6R manipulators~\cite{garnayak2019kinematics} capable of full six-dimensional (6-D) pose control of an end-effector, to contemporary dexterous robotic hands with on the order of 20 DoF~\cite{bicchi2000hands} and humanoid platforms with dozens of DoF~\cite{kaneko2002design}. These designs highlight the study of redundant robots in which number of joint DoF is higher than the dimensions required in the workspace, which leads to infinite solutions for one fixed target pose. The redundant dimension is frequently higher than 1-D when considering multi-fingered hands and humanoid robots' motion.

Burdick first proposed the term ``self-motion manifold (SMM)" that describes the geometric structure of an infinite number of solutions for redundant revolute manipulators~\cite{burdick1989inverse}. SMM offers a lens for computing global optimal redundancy-resolution strategies for various objectives, providing a way to escape local minima and avoid deadlocks with minimal efforts~\cite{fabregat2025topological}. However, SMM computation is fundamentally challenging because its non-Euclidean structure prevents straightforward global parameterization and becomes increasingly difficult to represent and explore as its dimensionality grows. Only simple manipulators' SMM such as 3R manipulators can be computed analytically~\cite{burdick1989characterization, guri2025ode}. Wu et al. try to use an evolution algorithm, cellular automata to search 1-D and 2-D SMMs~\cite{wu2023novel}. The work~\cite{guri2025ode} proposes a computation algorithm for 1-D SMM based on ordinary differential equation (ODE) methods. Later, Clark et al. first show a learning-based method to approximate the 1-D SMMs by learning the coefficients of Fourier series which represent a class of homotopic SMMs~\cite{clark2025learning}. The aforementioned methods~\cite{guri2025ode, clark2025learning} focus on 1-D SMMs and cannot be generalized to high dimensional SMMs trivially. The work~\cite{wu2023novel} tests the method for $r = 2$ but it is highly computationally expensive due to comprehensive search and clustering.

\begin{figure}[t]
    \centering
    \includegraphics[width=\linewidth]{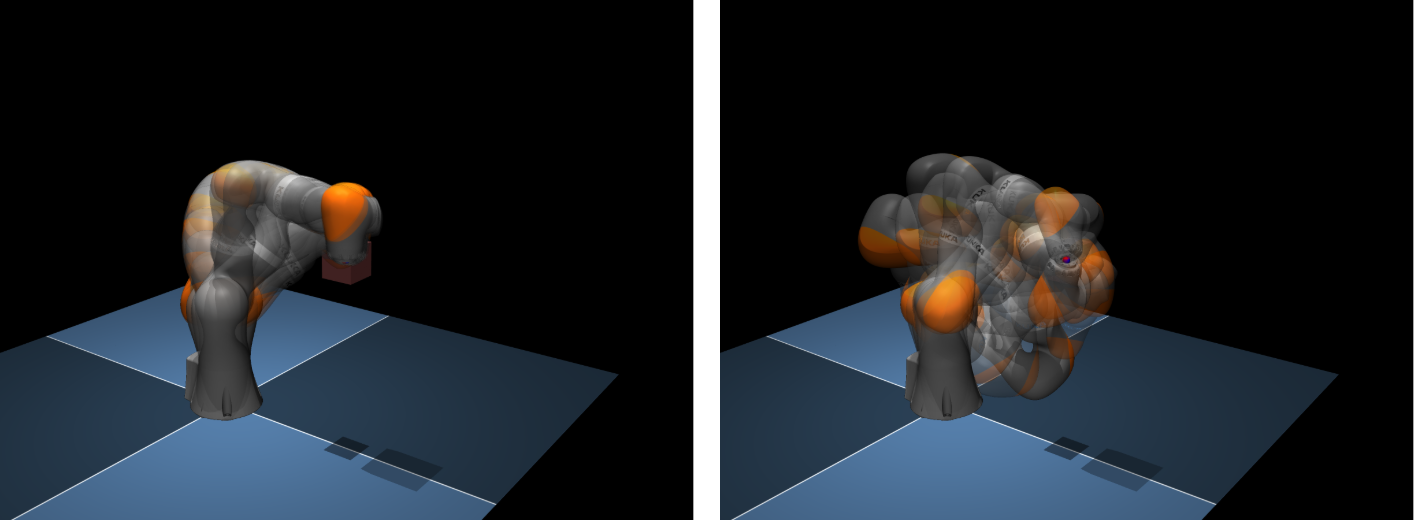}
    \caption{Approximation of self-motion manifolds (SMMs) using flow-matching models. For a 7-DoF KUKA iiwa robotic arm, a 6D full-pose target yields 1D SMMs (left), whereas a 3D position-only target yields 4D SMMs (right).}
    \label{fig: ghost demos}
\end{figure}

In this work, we adopt a probabilistic perspective and formulate SMM approximation as learning the conditional distribution $p(\mathbf{q} \mid \mathbf{x})$, where $\mathbf{q}$ denotes the robot configuration and $\mathbf{x}$ denotes the desired end-effector pose. Deep generative models have demonstrated strong expressive capacity for approximating complex and multimodal distributions, particularly in image and video synthesis~\cite{peebles2023scalable,brooks2024video}. Motivated by this capability, we learn conditional generative models whose samples represent the geometric structure of SMMs. The resulting SMM components can then be identified by clustering generated configurations under appropriate metrics.

Since the Fourier-based representation outperforms the grid-search method of Wu et al.~\cite{wu2023novel}, as reported by Clark et al.~\cite{clark2025learning}, we primarily compare our approach with the Fourier-based method and the ODE-based method of Guri and Kantor~\cite{guri2025ode,clark2025learning} for approximating one-dimensional SMMs. Unlike these approaches, our formulation naturally extends to higher-dimensional SMMs, including 4-D manifolds. To the best of our knowledge, the highest redundancy order addressed in prior SMM computation methods is $r=2$~\cite{wu2023novel}. In summary, our main contributions are as follows:
\begin{itemize}
    \item The first generative approach for SMM approximation by learning the conditional distribution $p(\mathbf{q} \mid \mathbf{x})$ and clustering samples drawn from this distribution to recover distinct SMM components.
    
    \item Our method generates configurations that satisfy joint limits by construction. In contrast, existing ODE- and Fourier-based representations do not directly accommodate joint limits, the enforcement of which can violate the periodic assumptions underlying these methods.
    
    \item The first method can be easily extended to approximate high dimensional SMMs accurately and efficiently which has not been achieved by existing works.

  \item The proposed framework is flexible, allowing the integration of alternative deep generative models and clustering algorithms in a plug-and-play manner.
\end{itemize}

\section{Related Works}
The main reason why SMM is an important object to investigate is that SMM includes all infinitely many solutions for a redundant manipulator given a fixed target pose, which is promising to give global motion plans especially in cluttered environments and fault-tolerant motions~\cite{almarkhi2019maximizing,bader2020hybrid}. 

The work~\cite{burdick1989inverse} proposed the term ``self-motion manifold" (SMM) to describe the infinite solutions for redundant revolute manipulators and gave the bounds of finite number of SMMs given a fixed target pose $\mathbf{x}$, namely, redundant planar, spherical, regional and spatial manipulators can respectively have as many as 2, 4, 8 and 16 distinct self-motions. 

\begin{figure*}[t]
    \centering
    \includegraphics[width=\linewidth]{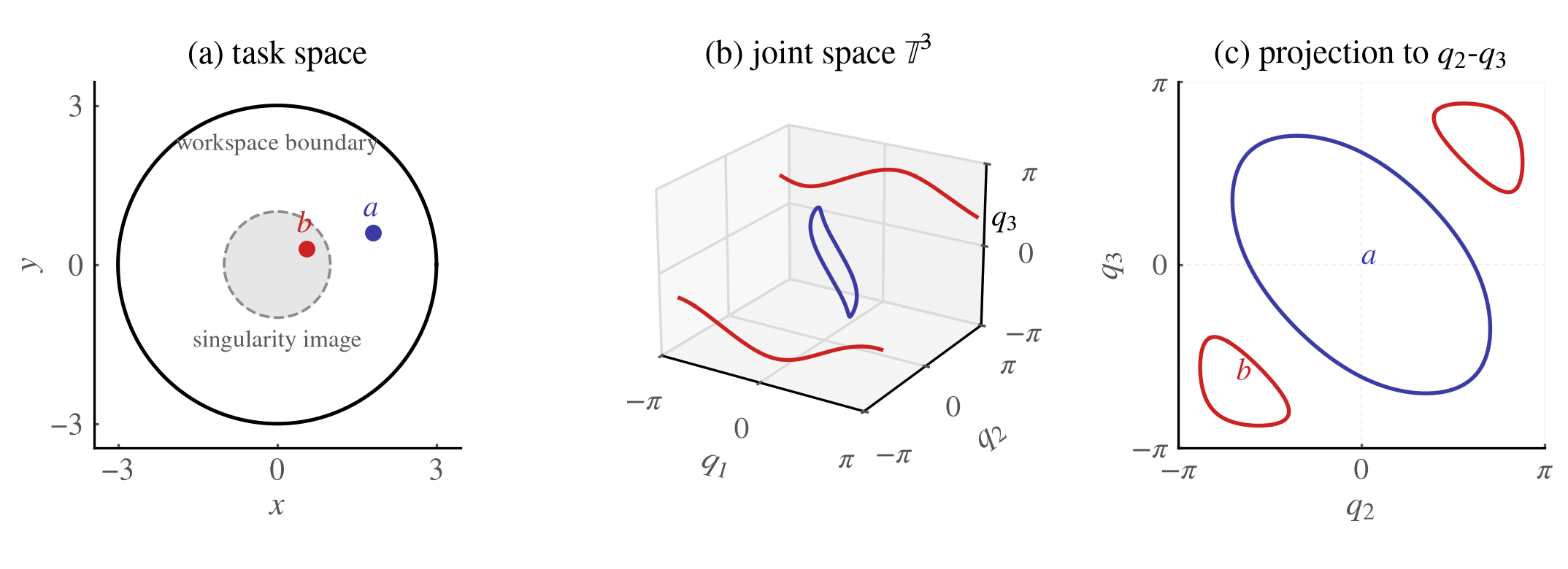}
    \caption{An example of self-motion manifolds of two positions in the workspace of 3R manipulator: (a) the task space of the 3R manipulator with unit length, the dashed circle is a singularity when two links are folded; (b) Two disjoint red curves are two sets of infinite solutions for the target point b and one closed curve is the set of infinite solutions of point a; (c) 3D projection onto the $q_2 -q_3$ plane.}
    \label{fig: example}
\end{figure*}

To understand SMM easily, Fig.~\ref{fig: example} shows the self-motion manifolds of a unit-length 
3R manipulator under a planar motion. Writing
$\rho$ for the base-to-target distance, the preimage $f^{-1}(\mathbf{x})$
comprises two disjoint closed curves for $\rho < 1$ and a single closed curve
for $1 < \rho < 3$, where the disc $\rho < 3$ is the reachable workspace. This simple example demonstrates the main features of SMMs:
\begin{itemize}
    \item SMMs are $r$ dimensional submanifold which is a 1-D curve in this case.
    \item SMMs branches vary based on the target pose in the workspace.
    \item SMMs  can sometimes be disjoint for the same target pose.
\end{itemize}

There are three main categories of SMMs computation: continuation methods, sampling-based methods and learning-based methods in the literature.

\paragraph{Continuation} This method is developed based on continuation in the Jacobian's null space which is also the tangent space of SMMs~\cite{burdick1989inverse}. The nullspace projection method~\cite{burdick1989inverse} determines the SMMs by iteratively moving in the Jacobian’s null space with a small step size to obtain a new configuration, and repeating this procedure until it traces back to the initial point: a closed curve. However, this method is sensitive near the singularity because of blow-up of pseudoinverse of the Jacobian matrix. An extension of this method~\cite{almarkhi2019maximizing} extends to high dimensional nullspaces. Another branch of computing SMMs is reducing redundant manipulators to the non-redundant manipulators and solving the inverse kinematics with a finite number of solutions~\cite{shimizu2008analytical}. In addition, a new ODE-based method is proposed to compute 1-D SMMs as an initial value problem, thereby enabling the use of standard
high-order explicit ODE integrators~\cite{guri2025ode}.

\paragraph{Sampling-based Method} This method involves two stages: (1) generate samples in $\mathbf{C}$-space. (2) clustering them to identify disjoint SMMs. The first such approach determines all SMMs corresponding to the homotopy classes of interest by randomly sampling joint configurations and then driving each sample to its nearest SMM using Jacobian-based inverse kinematics~\cite{rice2020multi}. The second approach divides the joint space into a grid and identifies the SMMs by exploring those grid cells that are intersected by the null space of their neighboring cells~\cite{wu2023novel}. 

\paragraph{Learning-based methods} With the growth of computational power, learning-based methods appear to approximate the SMMs with the expressive power of neural networks. Recently the study~\cite{clark2025learning} has integrated the classification capabilities of neural networks with a cellular automata search. In this framework, neural networks are employed both to identify the homotopy class of SMMs and to regress the Fourier-series coefficients that provide a closed-form representation of 1-D SMMs as curves.

However, existing methods either incur prohibitive computational cost, as in cellular-automata search~\cite{wu2023novel,clark2025learning}, or do not extend readily to the computation of high-dimensional SMMs~\cite{guri2025ode}. The difficulty compounds with the degree of redundancy: once the SMM is a surface or a higher-dimensional variety rather than a curve, exhaustive tracing of the full manifold becomes intractable.

In this work we instead adopt a probabilistic view of SMMs based on the sampling-based method principles. Not like existing sampling-based methods in which they generate the samples randomly or calculated by IK solvers, we treat the SMM as the support of the conditional posterior over configurations given a target pose, so that tracing the manifold reduces to sampling from a learned prior and recovering its disjoint components corresponds to clustering those samples. Deep generative models, especially the flow-matching approach~\cite{lipman2022flow} adopted here for its simplicity and greater stability relative to diffusion, fit this setting well: they are trained via direct regression, generate samples using a fixed number of network evaluations that does not depend on the manifold dimension, and scale readily to high-dimensional configuration spaces.

\section{Background}
\label{sec: bg}
\subsection{Configuration Space and Workspace}
For robotic manipulators, there are two distinct spaces: configuration space and workspace. 

Configuration space $\mathbf{C}$ describes the space of joint configurations. Each joint provides one DoF to the space. Formally, let $\mathbf{q}$ be the vector of joint variables,
\[
\mathbf{q} \in \mathbf{C}
\]
where $\mathbf{q} = \{q_1, q_2, \cdots, q_n\}$. Considering revolute manipulators whose joints just provide rotation, the geometric structure of $\mathbf{C}$ is a well-defined torus manifold such that,
\[
\mathbf{C} := S^1 \times S^1 \times \cdots \times S^1 := \mathbf{T}^n
\]

Workspace $\mathbf{W}$ characterizes the end-effector pose. Considering rigid motion, end-effector poses belong to special Euclidean group $SE(3)$ represented as a product of matrix exponential~\cite{selig2005geometric}. Formally, the pose $\mathbf{x}$ can be formulated as,
\[
\mathbf{x} \in \mathbf{W} \subseteq \textbf{SE(3)}
\]

\subsection{Redundant Manipulators and Self-motion Manifold}
A manipulator is redundant with respect to a task when the dimension of its
configuration space $\mathbf{C}$ exceeds that of the task space $\mathbf{W}$,
\[
r = n - m > 0
\]
where $r$ denotes the degree of redundancy, $n$ is the dimension of $\mathbf{C}$-space and $m$ is the dimension of $\mathbf{W}$-space. Redundancy is thus a property of the
manipulator--task pair rather than of the manipulator alone.

Let $f:\mathbf{C}\to\mathbf{W}$ denote the forward kinematics map. For a
reachable pose $\mathbf{x}\in\mathbf{W}$, the inverse kinematics solution set is
the preimage $f^{-1}(\mathbf{x})$. If $\mathbf{x}$ is a regular value of $f$,
equivalently, $\operatorname{rank}J(\mathbf{q})=m$ for every
$\mathbf{q}\in f^{-1}(\mathbf{x})$, the preimage theorem guarantees that
$f^{-1}(\mathbf{x})$ is a smooth $r$-dimensional submanifold of $\mathbf{C}$~\cite{lee2013smooth}. Each connected component of this preimage is termed a SMM: the manipulator may move continuously within a component while the
end-effector pose remains fixed. Burdick~\cite{burdick1989inverse} showed that
the preimage is in general disconnected, so that $f^{-1}(\mathbf{x})$ is a finite
disjoint union of self-motion manifolds,
\begin{equation}
    f^{-1}(\mathbf{x}) = \bigsqcup_{k=1}^{K(\mathbf{x})} \mathbf{M}_k(\mathbf{x}),
\end{equation}
where each $\mathbf{M}_k(\mathbf{x})$ is a connected $r$-dimensional manifold and
the number of components $K(\mathbf{x})$ depends on the pose.

\subsection{Ordinary differential equation for SMM Computation}
Guri et al.~\cite{guri2025ode} frame the computation of one-dimensional
self-motion manifolds as an initial value problem and solve it by standard high-order explicit ODE integrators. The starting point is the observation that a $k=1$ SMM is a curve in configuration space whose tangent is everywhere aligned with the (one-dimensional) kernel of the manipulator Jacobian, giving the unit internal-motion direction
\begin{equation}
    \hat{\mathbf{n}}(\mathbf{q}) = \frac{\ker(\mathbf{J}(\mathbf{q}))}{\lVert \ker(\mathbf{J}(\mathbf{q})) \rVert} \in \mathbb{R}^n .
\end{equation}
The naive autonomous ODE $\dot{\mathbf{q}} = \hat{\mathbf{n}}(\mathbf{q})$ is, however, impractical: the sign of $\hat{\mathbf{n}}(\mathbf{q})$ returned by a numerical kernel routine is ambiguous and may flip between successive samples, so the authors introduce a directionally regularized field that keeps the velocity aligned with a reference heading $\hat{\mathbf{n}}_{\mathrm{REF}}$,
\begin{equation}
    \dot{\mathbf{q}} = g(\mathbf{q}; \hat{\mathbf{n}}_{\mathrm{REF}}) =
    \begin{cases}
        \hat{\mathbf{n}}(\mathbf{q}), & \langle \hat{\mathbf{n}}(\mathbf{q}), \hat{\mathbf{n}}_{\mathrm{REF}} \rangle > 0, \\
        -\hat{\mathbf{n}}(\mathbf{q}), & \text{otherwise,}
    \end{cases}
\end{equation}
with $\hat{\mathbf{n}}_{\mathrm{REF}}$ carried over from the previous step. This regularized autonomous problem, termed the SMM initial value problem (SMM-IVP), is integrated with a fixed-step RK5 scheme ($h=0.05$) from an initial configuration obtained by any unconstrained IK solver, and is terminated once the trace returns to within one step of that configuration. Because the ODE is autonomous, the step size directly governs the arc length advanced per step. Notably, unlike the traditional trace update~\cite{burdick1989inverse}, this formulation requires no separate end-effector error-correction term, since the solver accuracy alone keeps the pose error less than $10^{-9}$.

However, the formulation is intrinsically 1-D. When each self-motion manifold
is a surface, the Jacobian kernel is a two dimensional plane rather than a line, so every
direction in the tangent plane is an admissible continuation and the sign
convention that resolves the ambiguity at $r = 1$ has no counterpart. Fixing a
direction still yields an integral curve, but a curve is a measure-zero subset
of a surface. Recovering
the manifold thus requires a chart of the domain and a rule for advancing over
it, not a single initial value problem.


\subsection{Representing SMMs as Fourier Series}
While the ODE and null-space approaches trace SMMs configuration-by-configuration at query time, \cite{clark2025learning} instead learn a closed-form approximation offline so that all SMM branches of an arbitrary workspace pose can be recovered in near-constant time. Their central idea is to represent each one-dimensional SMM branch, computed once by the null-space projection method as a sequence of discrete joint configurations, in the frequency domain via the discrete Fourier transform,
\begin{equation}
    \boldsymbol{\Theta}^{(k)}_{\mathbf{x}} = \sum_{i=1}^{N} \mathbf{T}\!\left(\boldsymbol{\theta}^{(i)}_{\mathbf{x}}\right) \exp\!\left(\frac{-j\,2\pi(k-1)(i-1)}{N}\right),
\end{equation}
so that a continuous branch is retrieved from a truncated set of coefficients by the inverse transform. Because a $k=1$ SMM can always be deformed into a unit circle and is therefore periodic~\cite{burdick1989inverse}, its high-frequency content is negligible, and retaining only the lowest $\sim\!40$ of $128$ components suffices to reconstruct the manifold with sub-milliradian error. To make the representation learnable, joint angles are first mapped onto the circle group $\mathbf{T}(\boldsymbol{\theta}) = \exp(j\boldsymbol{\theta})$ (respecting the $2\pi$ periodicity of revolute joints), and the intrinsic ambiguity in the starting configuration and traversal direction is resolved by fixing the phase of the first positive frequency component via the shift theorem and by selecting a consistent tangent through a closed-form null-vector sign convention.

However, the representation presumes a domain. At $r = 1$ this is available directly: every self-motion manifold is a closed curve, hence diffeomorphic to
$S^1$, so $\{e^{jks}\}$ is canonical and the residual ambiguity in starting
configuration and traversal direction is discrete and removable. At $r = 2$ no
such basis exists. The diffeomorphism type of a compact surface is not fixed a
priori and may vary with $\mathbf{x}$, and even granting a torus, a parameterization from
$\mathbb{T}^2 \to \mathbf{M}_k$ remains arbitrary: reparametrizing the domain
leaves the manifold unchanged but yields entirely different coefficients. The regression target is
therefore ill-posed, and the coefficient count grows as $(2K+1)^{r}$.

In addition, both methods ignore the joint limits which can break the periodicity and the property of being diffeomorphic to
$S^1$. A post-processing is required when these methods are applied to real robotic system.

\section{Flow-matching Models}
\label{sec:flow-matching}

Flow-matching model learns a
continuous-time transport from a tractable source density $p_0$ to a data density
$p_1$. Let $(p_t)_{t\in[0,1]}$ be a probability path and $u_t$ a time-dependent
velocity field generating it, i.e. satisfying the continuity equation
\begin{equation}
    \partial_t p_t + \nabla\cdot\!\left(p_t\, u_t\right) = 0 .
    \label{eq:continuity}
\end{equation}
Rather than fitting $u_t$ directly, which is intractable because $p_t$ is
unknown, one prescribes a conditional interpolant between a source sample
$\mathbf{q}_0\sim p_0=\mathbf{N}(\mathbf{0},\mathbf{I})$ and a data sample
$\mathbf{q}_1\sim p_1$. Under the linear (optimal-transport) interpolant,
\begin{equation}
    \mathbf{q}_t = (1-t)\,\mathbf{q}_0 + t\,\mathbf{q}_1 ,
    \qquad
    \dot{\mathbf{q}}_t = \mathbf{q}_1 - \mathbf{q}_0 ,
    \label{eq:interpolant}
\end{equation}
and the velocity field is regressed onto this conditional target,
\begin{equation}
    \mathbf{L}_{\mathrm{CFM}}(\theta) =
    \mathbb{E}_{t\sim\mathbf{U}[0,1],\;\mathbf{q}_0\sim p_0,\;\mathbf{q}_1\sim p_1}
    \left\|\, v_\theta(\mathbf{q}_t,t) - (\mathbf{q}_1 - \mathbf{q}_0) \,\right\|^2 .
    \label{eq:cfm-loss}
\end{equation}
The minimizer of~\eqref{eq:cfm-loss} is the conditional expectation
$u_t(\mathbf{q}) = \mathbb{E}\!\left[\dot{\mathbf{q}}_t \mid \mathbf{q}_t =
\mathbf{q}\right]$, which is precisely the marginal field generating $p_t$;
training therefore requires no simulation of the ODE and no divergence terms.
Sampling integrates the learned field forward,
\begin{equation}
    \frac{\mathrm{d}\mathbf{q}(t)}{\mathrm{d}t} = v_\theta\!\left(\mathbf{q}(t),t\right),
    \qquad \mathbf{q}(0)\sim p_0 ,
    \label{eq:ode-sampling}
\end{equation}
so that $\mathbf{q}(1)\sim p_1$, at a cost of $N$ network evaluations for an
$N$-step solver, independent of the dimension of $p_1$.

\subsection{Conditional Flow Matching Model for SMM}

We condition the velocity field on the target pose, $v_\theta(\mathbf{q},t,
\mathbf{x})$, and train it to transport $p_0$ to the IK posterior
$p(\mathbf{q}\mid\mathbf{x})$. Training pairs are obtained in the forward
direction, which requires no IK solver: configurations are drawn from a prior
$\rho$ over $\mathbf{C}$ and labeled by the FK,
\begin{equation}
    \mathbf{D} = \left\{ \left(\mathbf{q}^{(i)},\, \mathbf{x}^{(i)} =
    f(\mathbf{q}^{(i)})\right) \right\}_{i=1}^{M},
    \qquad \mathbf{q}^{(i)}\sim\rho ,
    \label{eq:dataset}
\end{equation}
and the objective becomes
\begin{equation}
    \mathbf{L}(\theta) =
    \mathbb{E}_{(\mathbf{q}_1,\mathbf{x})\sim\mathbf{D},\;t,\;\mathbf{q}_0}
    \left\|\, v_\theta(\mathbf{q}_t,t,\mathbf{x}) - (\mathbf{q}_1 - \mathbf{q}_0)
    \,\right\|^2 .
    \label{eq:cik-loss}
\end{equation}
Because every label in~\eqref{eq:dataset} is exact, the training distribution is
supported exactly on the fibers of $f$, and the posterior the model targets
satisfies
\begin{equation}
    \operatorname{supp} p(\cdot\mid\mathbf{x}) = f^{-1}(\mathbf{x}) =
    \bigsqcup_{k=1}^{K(\mathbf{x})} \mathbf{M}_k(\mathbf{x}) .
    \label{eq:support}
\end{equation}
Equation~\eqref{eq:support} is the link we exploit: the geometric object of
interest is the support of a distribution we can sample from, and its connected
components in which the self-motion manifolds are the modes of that distribution in
the sense of disjoint support, not of local density maxima.

\subsection{Recovering SMMs by sampling and clustering}

Given a query pose $\mathbf{x}$, we draw $N$ independent samples
$\{\mathbf{q}^{(j)}\}_{j=1}^{N}$ by integrating~\eqref{eq:ode-sampling} in
parallel, and retain those consistent with the query,
\begin{equation}
    \mathbf{S}_\delta(\mathbf{x}) = \left\{ \mathbf{q}^{(j)} :
    d_{\mathbf{W}}\!\left(f(\mathbf{q}^{(j)}),\mathbf{x}\right) \le \delta
    \right\} ,
    \label{eq:filter}
\end{equation}
a filter that is essentially free since it evaluates only FK. We can reduce the FK error further when the residual exceeds the tolerance required by the application, it can be
reduced by a Jacobian-based inverse kinematics update,
\begin{equation}
    \Delta \theta = \mathbf{J}^{\dagger} \Delta \mathbf{x}_{e},
\end{equation}
where $\mathbf{J}^{\dagger}$ denotes the Moore--Penrose pseudoinverse of the
manipulator Jacobian evaluated at the current configuration. This correction is
inexpensive and standard. The work~\cite{clark2025learning} also follows the same process.

The surviving samples lie in a $\delta$-tube about $f^{-1}(\mathbf{x})$. Since
distinct components are separated by a positive gap
$\gamma(\mathbf{x}) = \min_{k\neq l}\operatorname{dist}(\mathbf{M}_k,
\mathbf{M}_l)$, any clustering at a radius $\epsilon$ with
$\delta \ll \epsilon < \gamma(\mathbf{x})$ partitions
$\mathbf{S}_\delta(\mathbf{x})$ into groups
$\{\widehat{\mathbf{M}}_k\}_{k=1}^{\widehat{K}}$ that recover the components of
the fibre; the number of clusters $\widehat{K}$ is an estimate of
$K(\mathbf{x})$ and is not specified in advance. Distances are measured with the
intrinsic metric of the configuration space $\mathbf{C}\cong\mathbb{T}^{n}$,
\begin{equation}
    d_{\mathbf{C}}(\mathbf{q},\mathbf{q}')^2 = \sum_{i=1}^{n}
    \big(\!\operatorname{wrap}(q_i - q_i')\big)^2 ,
    \label{eq:metric}
\end{equation}
where $\operatorname{wrap}(\cdot)$ maps each joint difference to $(-\pi,\pi]$
ignoring the periodicity of the joint variables would split single components at
the coordinate cut. Tracing an SMM thus reduces to one batched forward pass, which can be performed
efficiently on a GPU, followed by a clustering step, replacing Jacobian
continuation or exhaustive search.

\subsection{Evaluation Metrics}
\label{sec:metrics}
We compare the three methods in terms of  FK error and inference time.

\paragraph{FK error.} For the configurations $\{\mathbf{q}_i\}_{i=1}^{N_{\mathrm{samp}}}$
returned by each method, we recompute FK and compare to the target.
On the 3R arm (position target $\mathbf{x}^\star$):
\begin{equation}
  e_{\mathrm{p}}(\mathbf{q}_i) = \lVert \mathbf{p}(\mathbf{q}_i) - \mathbf{x}^\star \rVert_2 / L
  \label{eq:pos-error}
\end{equation}
where $L$ is the maximum workspace length. On the 7R arm (target pose $(R^\star, \mathbf{p}^\star)$):
\begin{align}
  e_{\mathrm{p}}(\mathbf{q}_i) &= \lVert \mathbf{p}(\mathbf{q}_i) - \mathbf{p}^\star \rVert_2 / L, \\
  e_{\mathrm{o}}(\mathbf{q}_i) &= \arccos\!\left( \tfrac{\operatorname{tr}(R(\mathbf{q}_i)^{\top} R^\star) - 1}{2} \right) / \pi, \\
  e_{\mathrm{overall}} &= \frac{1}{2}(e_{\mathrm{p}} + e_{\mathrm{o}}) 
  \label{eq:pose-error}
\end{align}
We report the mean of all tested samples.

\paragraph{Efficiency} We assess algorithmic efficiency by reporting (i) training time, including data generation, in seconds for learning-based methods, and (ii) inference time, in milliseconds, for all methods.

\section{Results}
We validate our proposed method on a planar 3R and 7R KUKA iiwa for full 6-D pose  and compare its performance with ODE and Fourier series representation about 1-D SMMs approximation as described in Sec.~\ref{sec: bg}. We then demonstrate that the flow-matching model can be easily extended to approximate 4-D SMMs, using a 7R manipulator performing 3D position task to highlight the advantages.

\subsection{Simulation Software}
In this work, we train the planar 3R robot in a simulation environment using ``\textit{robotics-toolbox-python}"~\cite{rtb} for simplicity. KUKA iiwa 14 is trained and visualized by Mujoco physics engine~\cite{todorov2012mujoco}. The DH parameters of these manipulator are listed in the Table~\ref{tab:dh-combined}. The hyperparameters used in the flow-matching model are listed in Table~\ref{tab:fm-hyperparams}. The hyperparameters used in Fourier model and ODE method are kept same with the original works and be listed in Appendix~\ref{sec: appendix}.

\begin{table}[t]
\centering
\caption{Standard Denavit--Hartenberg parameters of the planar 3R
manipulator and KUKA LBR iiwa14.}
\label{tab:dh-combined}
\renewcommand{\arraystretch}{1.15}
\begin{tabular}{lccccc}
\toprule
\textbf{Robot} & \textbf{Link $i$} & \textbf{$\theta_i$}
& \textbf{$d_i$ (m)} & \textbf{$a_i$ (m)}
& \textbf{$\alpha_i$} \\
\midrule

\multirow{3}{*}{Planar 3R}
& 1 & $q_1$ & 0 & 1 & $0^\circ$ \\
& 2 & $q_2$ & 0 & 1 & $0^\circ$ \\
& 3 & $q_3$ & 0 & 1 & $0^\circ$ \\

\midrule

\multirow{7}{*}{KUKA iiwa14}
& 1 & $q_1$ & 0.360 & 0 & $-90^\circ$ \\
& 2 & $q_2$ & 0     & 0 & $90^\circ$ \\
& 3 & $q_3$ & 0.420 & 0 & $90^\circ$ \\
& 4 & $q_4$ & 0     & 0 & $-90^\circ$ \\
& 5 & $q_5$ & 0.400 & 0 & $-90^\circ$ \\
& 6 & $q_6$ & 0     & 0 & $90^\circ$ \\
& 7 & $q_7$ & 0.081 & 0 & $0^\circ$ \\

\bottomrule
\end{tabular}
\end{table}

\begin{table}[t]
\centering
\caption{Flow-matching model architecture and training hyperparameters
for the planar 3R and KUKA LBR iiwa14 manipulators.}
\label{tab:fm-hyperparams}
\renewcommand{\arraystretch}{1.08}
\begin{tabular}{lll}
\toprule
\textbf{Component} & \textbf{3R arm} & \textbf{KUKA iiwa14} \\
\midrule
\multicolumn{3}{l}{\textit{Architecture}} \\
Joint-state dimension $q_{\mathrm{dim}}$
    & 3 & 7 \\
Conditioning dimension $x_{\mathrm{dim}}$
    & 2 & 3 (position) / 9 (pose) \\
Time embedding dimension
    & \multicolumn{2}{c}{64 (sinusoidal, frequency 1--1000)} \\
Hidden width
    & 256 & 512 \\
Hidden layers
    & 4 & 5 \\
Activation
    & \multicolumn{2}{c}{SiLU} \\
Input dimension
    & 69 & 74 (position) / 80 (pose) \\
Output dimension
    & 3 & 7 \\
\midrule
\multicolumn{3}{l}{\textit{Optimization}} \\
Optimizer
    & \multicolumn{2}{c}{AdamW} \\
Learning rate
    & \multicolumn{2}{c}{$1\times10^{-3}$} \\
Weight decay
    & \multicolumn{2}{c}{$1\times10^{-5}$} \\
Learning-rate schedule
    & \multicolumn{2}{c}{Cosine annealing} \\
Batch size
    & 512 & 1024 \\
Training epochs
    & \multicolumn{2}{c}{200} \\
Train/validation split
    & \multicolumn{2}{c}{90\% / 10\%} \\
Model selection
    & \multicolumn{2}{c}{Best validation CFM loss} \\
\midrule
\multicolumn{3}{l}{\textit{Sampling and inference}} \\
Sampling ODE integrator
    & \multicolumn{2}{c}{Explicit Euler} \\
Sampling steps
    & \multicolumn{2}{c}{100} \\
Prior distribution $q_0$
    & \multicolumn{2}{c}{$\mathbf{N}(\mathbf{0},I)$} \\
Random seed
    & \multicolumn{2}{c}{42} \\
\bottomrule
\end{tabular}
\end{table}

\subsection{1-D SMM Approximation for Planar 3R}

\begin{figure}
    \centering
    \includegraphics[width=\linewidth]{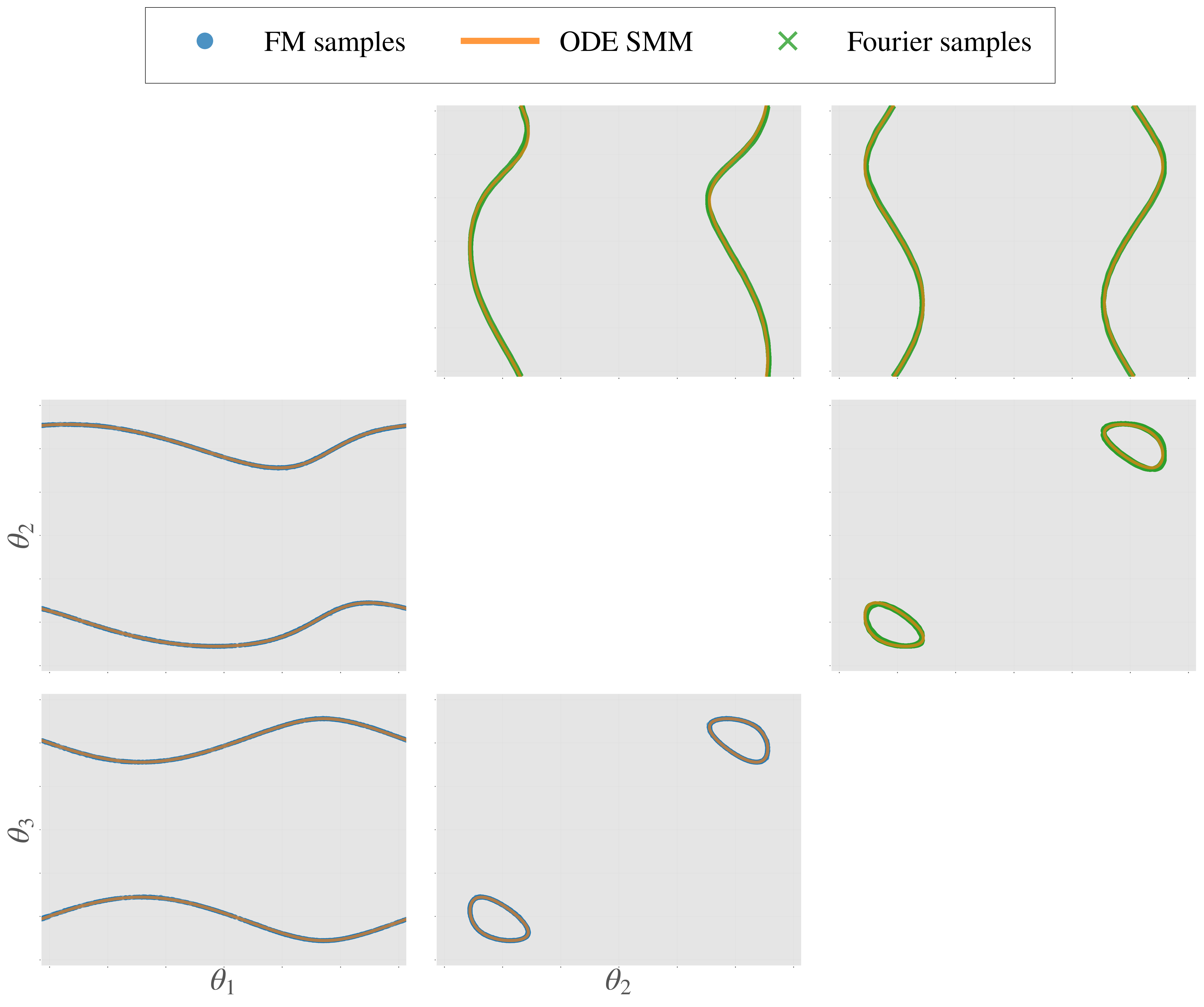}
    \caption{Pairwise joint-space comparisons for the Planar3R robot. The lower triangular panels compare the ODE and flow-matching models, while the upper triangular panels compare the ODE and Fourier models.}
    \label{fig: results for planar3R}
\end{figure}

For the planar 3R manipulator with unit link lengths and a position-only task, the self-motion manifolds are 1-D curves that can be enumerated in closed form. As shown in Table~\ref{tab:evaluation-metrics}, the raw FM samples achieve lower FK error than the Fourier model, although both are less accurate than the ODE baseline, which serves as the reference. The Fourier also provides the fastest inference. For this simple manipulator, Fig.~\ref{fig: results for planar3R} shows that all three methods closely overlap the same self-motion curves. The lower-triangle panels compare the ODE and flow-matching models and  the upper-triangular panels compare the ODE and Fourier models.

\begin{table*}[t]
    \centering
    \caption{Evaluation metrics for the different methods and 3 iterations Jacobian-corrected variants on Planar3R and KUKA iiwa. Position and orientation errors are reported as percentages.}
    \label{tab:evaluation-metrics}
    \begin{tabular}{llcccc}
        \toprule
        Robot & Method & $e_p$ (\%) & $e_o$ (\%) & $e$ (\%) & Inference time (ms) \\
        \midrule

        \multirow{5}{*}{Planar3R}
        & ODE
        & 2.95e-09 & 0 & 2.95e-09 & 166.03 \\
        & Fourier
        & 5.13e-01 & 0 & 5.13e-01 & \textbf{16.77} \\
        & Fourier + Jacobian correction
        & 5.29e-13 & 0 & 5.29e-13 & 29.70 \\
        & FM
        & 1.44e-01 & 0 & 1.44e-01 & 124.91 \\
        & FM + Jacobian correction
        & \textbf{1.56e-14}
        & \textbf{0}
        & \textbf{1.56e-14}
        & 126.46 \\

        \midrule

        \multirow{5}{*}{KUKA iiwa}
        & ODE
        & \textbf{2.57e-10} & 2.37e-06 & 1.18e-06 & 8209.95 \\
        & Fourier
        & 1.99e+01 & 4.70e+01 & 3.35e+01 & \textbf{39.93} \\
        & Fourier + Jacobian correction
        & 5.52e+00 & 8.46e+00 & 6.99e+00 & 144.09 \\
        & FM
        & 1.07e+00 & 1.70e+00 & 1.39e+00 & 296.89 \\
        & FM + Jacobian correction
        & 2.88e-10
        & \textbf{0}
        & \textbf{1.44e-10}
        & 374.88 \\

        \bottomrule
    \end{tabular}
\end{table*}

\subsection{1-D SMM Approximation for 7-DoF Manipulators}

\begin{figure}
    \centering
    \includegraphics[width=\linewidth]{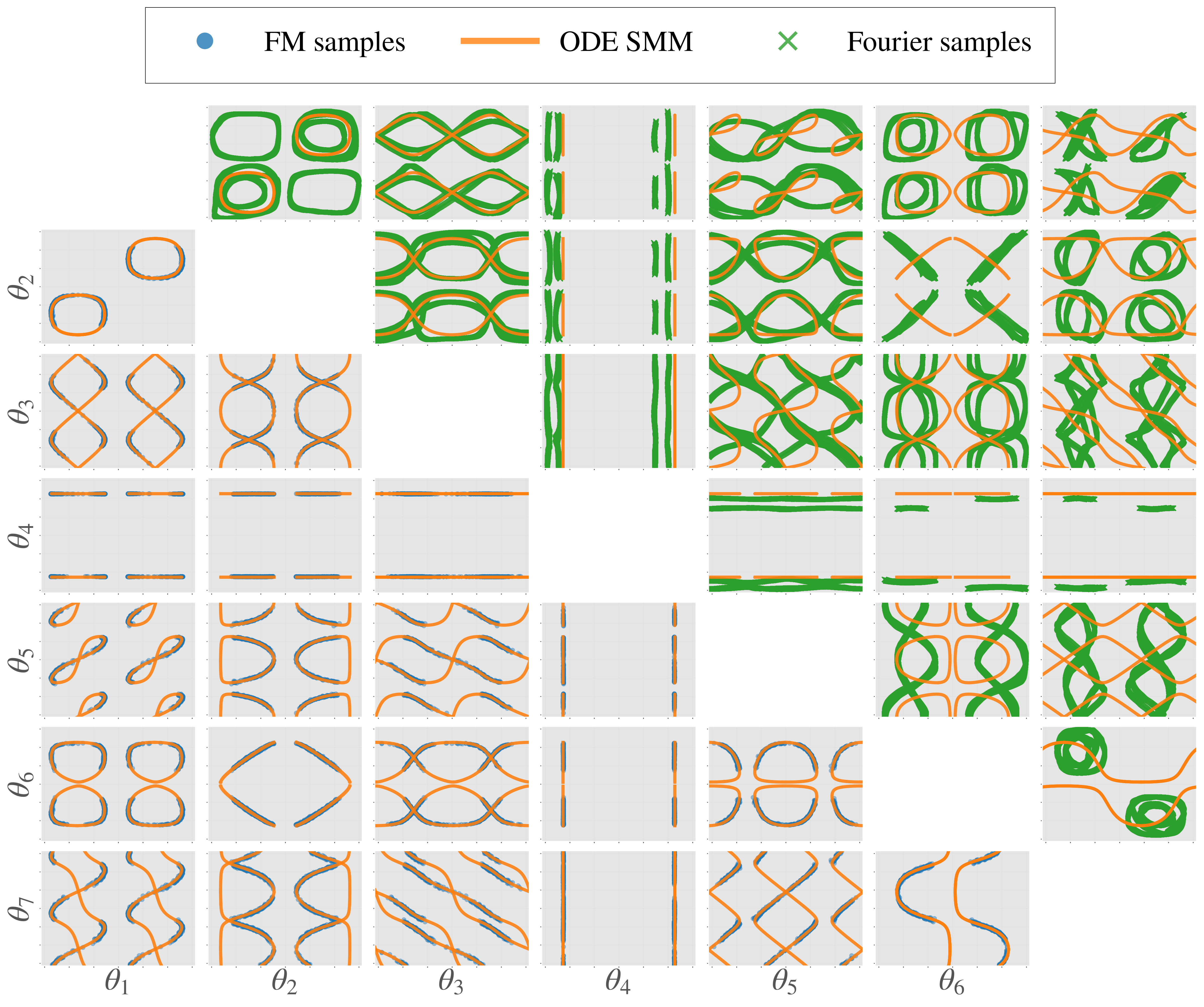}
    \caption{Pairwise joint-space comparisons for the KUKA iiwa. The lower triangular panels compare the ODE and flow-matching models, while the upper triangular panels compare the ODE and Fourier models.}
    \label{fig: results for iiwa}
\end{figure}

For the 7R manipulator, we consider the complete 6-D pose constraint, for which the self-motion manifolds are 1-D curves ($r=1$). The ODE and Fourier baselines are therefore evaluated only in this setting, since both methods are specifically formulated to represent self-motion manifolds as curves. As reported in Table~\ref{tab:evaluation-metrics}, the uncorrected FM model achieves a total FK error of $1.39\%$ on the KUKA iiwa, substantially lower than the Fourier model ($3.35\times10^{1}\%$). FM is also considerably faster than the ODE baseline, requiring $296.89$~ms compared with $8209.95$~ms. Applying three iterations Jacobian correction further reduces the FM error to $1.44\times10^{-10}\%$, approaching the ODE accuracy ($1.18\times10^{-6}\%$) while retaining a substantially lower inference time. The corrected FM model also significantly outperforms the corrected Fourier model, whose total error remains $6.99\%$. Qualitatively, Fig.~\ref{fig: results for iiwa} shows that FM samples closely follow the ODE-generated 1-D self-motion curve across the joint-pair projections. In contrast, Fourier samples exhibit substantial deviations and reproduce incorrect curve geometries in several projections. The FM samples cover only feasible portions of the complete curve because joint-limit constraints partition the unconstrained self-motion manifold into multiple segments. Visually, Fig.~\ref{fig: ghost demos} (left) shows that diverse valid solutions enable robot self-move without changing the end-effector's 6-D pose.

\subsection{High-Dimensional SMM Approximation}

Both baselines are inherently 1-D. The ODE method traces a self-motion component by integrating a null-space direction, which is unique up to sign, while the Fourier method represents the manifold using harmonic coefficients over $S^{1}$, the canonical domain for 1-D self-motion manifolds. Their training targets are also generated using the same 1-D continuation procedure. Consequently, neither method extends naturally to self-motion manifolds with redundancy $r=n-m\geq2$. In contrast, our generative formulation models the conditional distribution $p(\mathbf{q} \mid \mathbf{x})$ directly in joint space, so the output dimension remains $n$, independently of the manifold dimension $r$.

\begin{figure}[h]
    \centering
    \includegraphics[width=\linewidth]{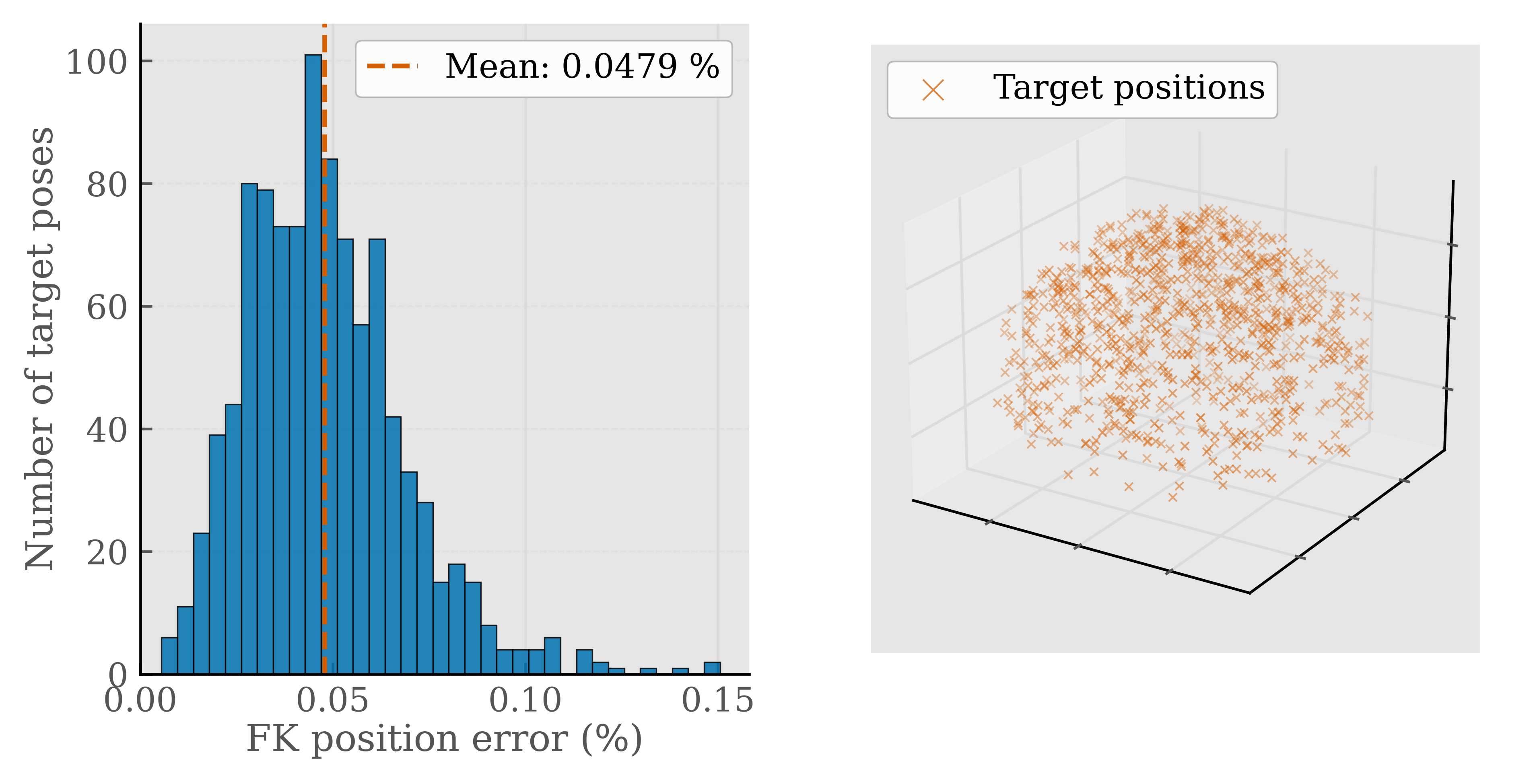}
    \caption{FK position error for approximated 4-D SMMs before Jacobian correction}
    \label{fig: FK error for 4-d smms 1}
\end{figure}

\begin{figure}[h]
    \centering
    \includegraphics[width=\linewidth]{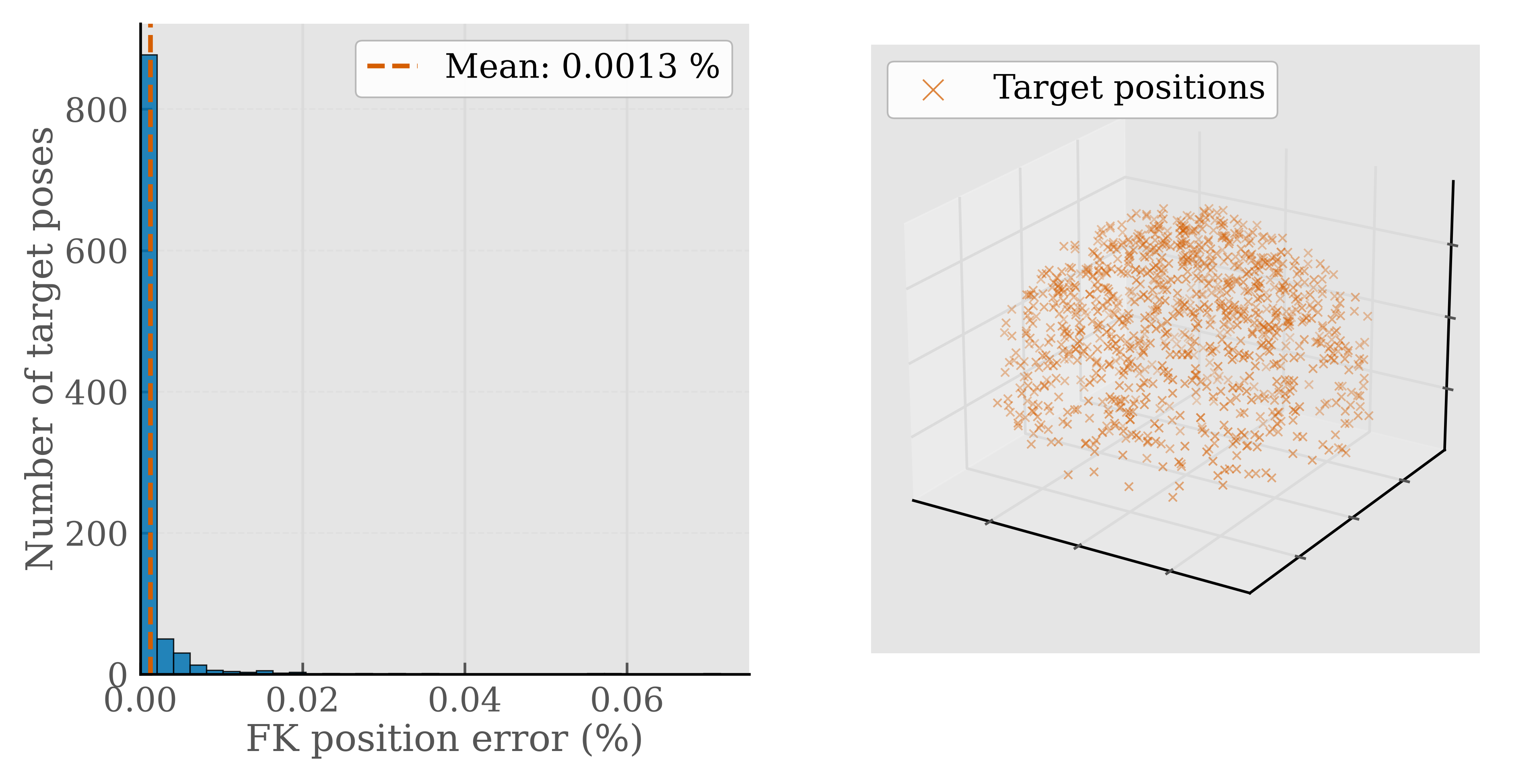}
    \caption{FK position error for approximated 4-D SMMs after Jacobian correction}
    \label{fig: FK error for 4-d smms 2}
\end{figure}

We demonstrate this capability using a 7R manipulator subject to a position-only task. Here, $n=7$, $m=3$, and therefore $r=4$; for each regular target position, the corresponding self-motion manifold is a smooth four-dimensional submanifold of $\mathbb{T}^{7}$. Given a target position $x\in\mathbb{R}^{3}$, the model generates joint configurations $q\sim p(q\mid x)$ using the same architecture, training objective, and sampling procedure as in the lower-dimensional tasks. Since a 4-D manifold cannot be visualized directly, we evaluate the generated configurations through their FK position errors and workspace coverage over 1,000 randomly selected test targets. As shown in Fig.~\ref{fig: FK error for 4-d smms 1}, the error histograms show mean FK errors of $0.0479\%$ before Jacobian correction and $0.0013\%$ after correction in Fig.~\ref{fig: FK error for 4-d smms 2}, relative to the maximum robot reach. The corresponding workspace plots show that the generated configurations reach the desired target positions. These results demonstrate that the model can generate multiple distinct valid joint configurations for the same Cartesian position and approximate high-dimensional self-motion manifolds without requiring a change in the model formulation. Qualitatively, Fig.~\ref{fig: ghost demos} (right) shows that the robot is able to reach the same position using configurations sampled from learned flow-matching model.

\subsection{Training Efficiency}
\begin{figure}
    \centering
    \includegraphics[width=\linewidth]{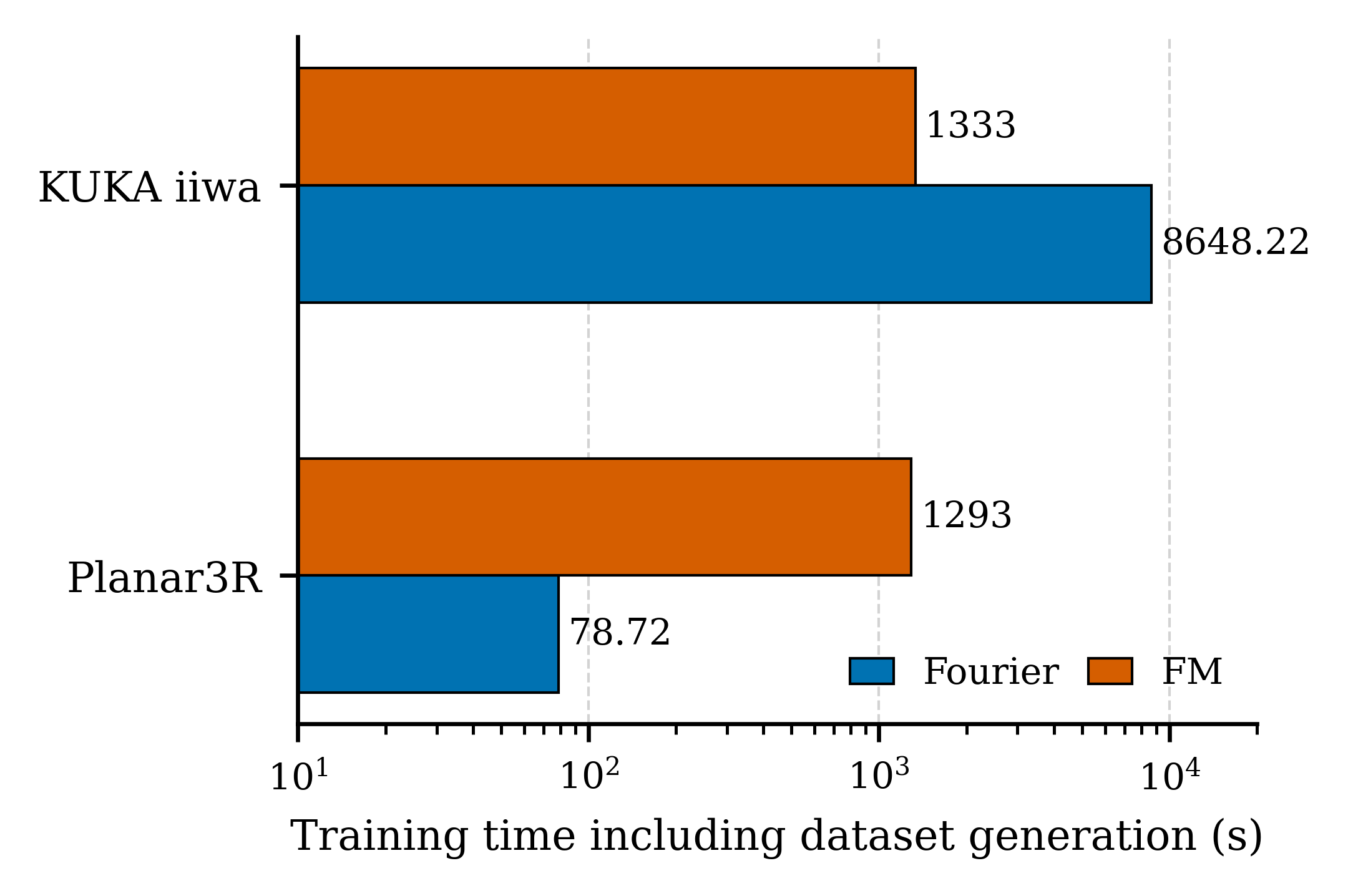}
    \caption{Training time, including dataset generation, for Fourier and flow-matching models across the Planar3R and KUKA iiwa robots.}
    \label{fig: training-time}
\end{figure}

The training efficiency of the flow-matching model exhibits limited sensitivity to the manipulator’s degrees of freedom (DoF), in contrast to the Fourier representation baseline. More generally, dataset size constitutes a primary bottleneck for generative-model training. As shown in Fig.~\ref{fig: training-time}, the Fourier representation method achieves substantially shorter training time in the low-DoF setting. However, its computational burden increases sharply as the manipulator complexity grows, reaching nearly 2 hours in the high-DoF case. Notably, this runtime corresponds to a manually optimized implementation of the original Fourier representation code; using the unmodified version would require approximately six times longer training time. Additional implementation details are provided in the Appendix. In comparison, on a dataset containing 2,000,000 samples, the flow-matching models train consistently within approximately 25 minutes.

\section{Discussion}

We presented a probabilistic framework for approximating SMMs by formulating the problem as learing a conditional distribution learning followed
by component clustering. The proposed approach complements existing geometric
and algebraic methods while providing a unified treatment of SMM approximation.

The framework offers several practical advantages. Samples drawn from the
learned conditional distribution $p(\mathbf{q} \mid \mathbf{x})$ are substantially more accurate
than the raw outputs of the Fourier-series representation, while inference is
faster than ODE-based continuation for the 7R manipulator. Training is also
simpler than the Fourier-series pipeline, which requires a prior workspace
clustering stage and two separate networks; in contrast, our approach trains a
single model end to end. Furthermore, the formulation extends naturally to
redundancy orders $r \geq 2$ without architectural changes, since it represents
a distribution supported on the manifold rather than an explicit
parametrization. The flow-matching model also supports large-batch sampling
via GPU parallelism, increasing the probability that at least one sample
satisfies the task constraints with only a small additional inference cost.

Another advantage is that the learned distribution is automatically restricted
to the joint-limit feasible subset of the self-motion manifold. By comparison,
continuation and Fourier-series methods represent the manifold independently
of joint limits, meaning that the recovered set is effectively a truncation of
the full manifold. Nevertheless, the proposed model remains tied to the task
specification: changing the dimension of $\mathbf{x}$ requires retraining, although the
underlying dataset of configuration--pose pairs can be reused.

More broadly, the increasing availability of GPU computing has shifted
SMM approximation from online local linearization based on Jacobians toward
offline precomputation and clustering followed by fast inference. This shift
opens several directions for future work. On the theoretical side, convergence
guarantees for component separation and the scaling of sample complexity with
the redundancy order $r$ remain to be established. On the application side,
the recovered components, or their homotopy classes, could be used to annotate
configuration space and provide global structural information for sampling-based and graph-based motion planners, which typically lack such guidance.

\addtolength{\textheight}{-12cm}   



\section*{APPENDIX}
\label{sec: appendix}
We adopt the hyperparameters of the Fourier-series representation~\cite{clark2025learning}
and the null-space ODE~\cite{guri2025ode} listed in table~\ref{tab:hyperparameters}. In addition, we
upgrade the Fourier implementation so that results for a 7-DoF arm can be
obtained within a reasonable time budget:
\begin{itemize}
    \item Batched forward kinematics.
    \item Faster workspace clustering for high-DoF arms, e.g.\ 7 DoF.
    \item Grid reduction by Fibonacci sampling in $SO(2)$ instead of uniform
          sampling.
    \item Support for the Franka Panda and KUKA iiwa\,14 robots.
\end{itemize}

\begin{table}[t]
\centering
\caption{Hyperparameters. Robot-specific entries are listed as
3R / Panda / iiwa\,14.}
\label{tab:hyperparameters}
\footnotesize
\setlength{\tabcolsep}{4pt}
\begin{tabular}{@{}ll@{}}
\toprule
Parameter & Value \\
\midrule
\multicolumn{2}{@{}l}{\textit{Fourier SMM --- grid}} \\
$\mathbf{x}$ range (m)                & $[-3,3]$ / $[-1,1]$ / $[-1,1]$ \\
$z$ range (m)                & -- / $[-0.5,1.3]$ / $[-0.6,1.31]$ \\
Position resolution (m)      & $0.005$ / $0.1$ / $0.1$ \\
Fibonacci directions, $SO(3)$ neighbors & $96$, $k=6$ (7R only) \\
\addlinespace[2pt]
\multicolumn{2}{@{}l}{\textit{Fourier SMM --- tracing}} \\
Samples per manifold, step   & $128$, $0.05$ \\
Max.\ iterations, IK seeds   & $1200$, $40$ \\
Branch threshold             & $0.15$ \\
Singularity threshold        & $5\times10^{-3}$ \\
Drift damping $\lambda$      & $10^{-3}$ \\
Max.\ manifolds per node     & $2$ / $16$ / $16$ \\
\addlinespace[2pt]
\multicolumn{2}{@{}l}{\textit{Fourier SMM --- network}} \\
Hidden layers                & $6\times150$, leaky ReLU \\
Epochs, batch size           & $1500$, $64$ \\
Learning rate, weight decay  & $10^{-3}$, $10^{-4}$ (AdamW) \\
FFT cutoff, min.\ cluster    & $24$, $8$ nodes \\
\addlinespace[2pt]
\multicolumn{2}{@{}l}{\textit{Null-space ODE}} \\
Integrator, step $h$         & Cash--Karp RK5, $0.05$ \\
Min.\ / max.\ steps          & $30$ / $20\,000$ \\
Closure tolerance            & $0.05$ \\
Singularity tolerance        & $\sigma_{\min}\!\le\!10^{-9}\max(1,\sigma_{\max})$ \\
IK attempts, iterations      & $150$, $100$ \\
IK tolerance, damping        & $10^{-10}$, $10^{-3}$ \\
Seed uniqueness tolerance    & $10^{-4}$ \\
\addlinespace[2pt]
\multicolumn{2}{@{}l}{\textit{Correction (both methods)}} \\
DLS steps, damping           & $3$, $\lambda=10^{-3}$ \\
\bottomrule
\end{tabular}
\end{table}

\bibliographystyle{IEEEtran}
\bibliography{references}

\end{document}